\documentclass[conference]{IEEEtran}
\IEEEoverridecommandlockouts
\usepackage{cite}
\usepackage{amsmath,amssymb,amsfonts}
\usepackage{algorithmic}
\usepackage{graphicx}
\usepackage{textcomp}
\usepackage{xcolor}
\usepackage{verbatim}
\usepackage{multicol}
\def\BibTeX{{\rm B\kern-.05em{\sc i\kern-.025em b}\kern-.08em
    T\kern-.1667em\lower.7ex\hbox{E}\kern-.125emX}}
    
\newtheorem{definition}{Definition}

\makeatletter
\def\ps@IEEEtitlepagestyle{
  \def\@oddfoot{\mycopyrightnotice. %
  }
  \def\@evenfoot{}
}

\makeatletter
\def\ps@IEEEtitlepagestyle{
  \def\@oddfoot{\mycopyrightnotice}
  \def\@evenfoot{}
}
\def\mycopyrightnotice{
  {\footnotesize
  \begin{minipage}{\textwidth}
  \centering
  ~\copyright~2020 IEEE. Personal use of this material is permitted. Permission from IEEE must be obtained for all other uses, in any current or future media, including reprinting/republishing this material for advertising or promotional purposes, creating new collective works, for resale or redistribution to servers or lists, or reuse of any copyrighted component of this work in other works.
  \end{minipage}
  }
}

\makeatother

\begin{document}

\title{A Cognitive Approach based on the Actionable Knowledge Graph for supporting Maintenance Operations}

\author{\IEEEauthorblockN{Giuseppe Fenza~\IEEEmembership{Member, IEEE}}
\IEEEauthorblockA{\textit{Department of Management \& } \\
\textit{Innovation Systems,} \\
University of Salerno\\
 Fisciano (SA), Italy\\
gfenza@unisa.it}
\and
\IEEEauthorblockN{Mariacristina Gallo~\IEEEmembership{Member, IEEE}}
\IEEEauthorblockA{\textit{Department of Management \&} \\
\textit{Innovation Systems,}\\
University of Salerno,\\
Fisciano (SA), Italy\\
mgallo@unisa.it}
\and
\IEEEauthorblockN{Vincenzo Loia~\IEEEmembership{Senior Member, IEEE}}
\IEEEauthorblockA{\textit{Department of Management \& } \\
\textit{Innovation Systems,}\\
University of Salerno, Fisciano (SA), Italy\\
loia@unisa.it}
\and
\IEEEauthorblockN{Domenico Marino}
\IEEEauthorblockA{\textit{Department of Management \& } \\
\textit{Innovation Systems,}\\
University of Salerno, Fisciano (SA), Italy\\
domenicomarino42@gmail.com}
\and
\IEEEauthorblockN{Francesco Orciuoli}
\IEEEauthorblockA{\textit{Department of Management \& } \\
\textit{Innovation Systems,}\\
University of Salerno, Fisciano (SA), Italy\\
forciuoli@unisa.it}
}
\maketitle


\begin{abstract}
In the era of Industry 4.0, cognitive computing and its enabling technologies (Artificial Intelligence, Machine Learning, etc.) allow to define systems able to support maintenance by providing relevant information, at the right time, retrieved from structured companies’ databases, and unstructured documents, like technical manuals, intervention reports, and so on. Moreover, contextual information plays a crucial role in tailoring the support both during the planning and the execution of interventions. Contextual information can be detected with the help of sensors, wearable devices, indoor and outdoor positioning systems, and object recognition capabilities (using fixed or wearable cameras), all of which can collect historical data for further analysis. 

In this work, we propose a cognitive system that learns from past interventions to generate contextual recommendations for improving maintenance practices in terms of time, budget, and scope. The system uses formal conceptual models, incremental learning, and ranking algorithms to accomplish these objectives.
\end{abstract}


\section{Introduction}
\label{sec:1}
Nowadays, in the industry sector, considerable attention is dedicated to support and foster innovation by using emergent technologies and frameworks \cite{18}. \emph{Industry 4.0} \cite{21}, also based on the paradigms of the Cyber-Physical Systems \cite{19}, leverages on the idea to embed intelligence into industrial products and systems in order to enable technologies for predicting product performance degradation, and autonomously manage
and optimize product service needs \cite{1}.
Moreover, in the area of service operations, the term \emph{Service 4.0} summarizes the idea that the current technological innovations could provide a great opportunity to define on-site, real-time supporting solutions for field service professionals in many areas. Thus, the main challenge for companies is to increase employees’ expertise and their ability to maintain and improve service quality~\cite{22}\cite{2}.

Under the context mentioned above, intelligent analytics and cyber-physical systems should be used in combination in order to realize new thinking of production management and factory transformation. Using appropriate sensor installations, various signals such as vibration, pressure, temperature, etc. can be extracted. Besides, historical data can be gathered for further data mining, in many cases sensorization and data analysis are not enough to detect faults or alarms and once they occur, an operator must fix them manually \cite{20}. Therefore, it is needed to manage, aggregate, and process huge volumes of heterogeneous data, i.e., Big Data~\cite{1}.

\begin{figure*}[ht]
\centering
\includegraphics[width=0.6\linewidth]{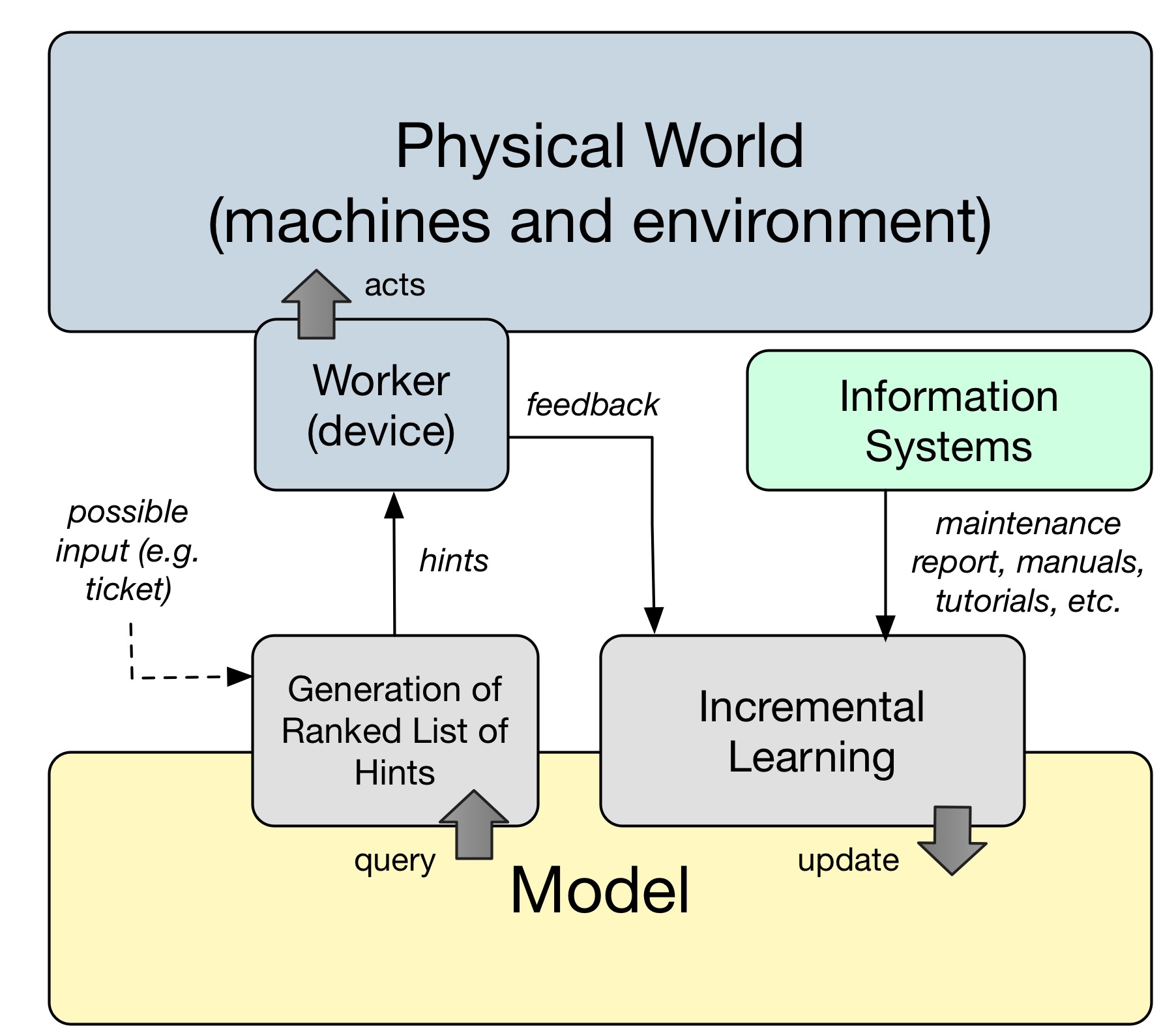}
\caption{Sketch of the proposed Cognitive System.}
\label{cogsys}
\end{figure*}

Cognitive Computing \cite{3, 13, 14} is a technology approach that enables humans to collaborate with machines to analyze all types of data, from structured data in databases to unstructured data in text, images, voice, sensors, and video. The synergy between humans and machines is represented by systems that operate at a different level than traditional IT systems because they analyze and learn from the aforementioned data. 

More in details, a cognitive system has three main principles: 
\begin{itemize}
\item \emph{Learn} - A cognitive system must learn, i.e., it leverages data to make inferences about a domain, a topic, a person, or an issue based on training and observations from all varieties, volumes, and velocity of data;
\item \emph{Model} - A cognitive system needs to create a model or representation of a domain (which includes internal and potentially external data) in order to enable the \emph{learning} principle and to understand the context of how the data fits into such a model; 
\item \emph{Generate hypotheses} - A cognitive system is probabilistic. Typically, it assumes that there is not a single correct answer. Therefore, a cognitive system is probabilistic, in the sense that it uses the data to train, test, or score a hypothesis.  
\end{itemize}

In this work, we propose a cognitive system (see Fig.~\ref{cogsys}) for supporting the maintenance operations of physical equipment. The idea is providing workers (on-the-job) with effective and efficient tools to make context-aware access (mainly in push logic as recommendations) to data (manuals, tutorials, procedures, historical data, etc.) while they are involved in real-world situations. This can mitigate the increasing complexity of industrial maintenance and repair tasks and reduce users’~cognitive load by providing support at any time by generating and delivering contextualized hints~\cite{13}.

The proposed system is based on the adoption of Formal Concept Analysis (FCA) to learn and evolve the model of the ecosystem in which the maintenance operations are applied by mining data traced during the execution of such operations (immediate feedback, intervention report, etc.), manuals, procedures, etc. In particular, the model is used by specific algorithms able to analyze the received trouble tickets, search over the model and generate ranked lists of situational recommendations to be provided to the workers involved in specific maintenance interventions on-the-field. The immediate feedback of workers to the received recommendations will refine the model. In this context, Augmented Reality (AR) could be adopted as a medium able to play both the role of displaying contextual useful information to the worker involved in the maintenance and the role of gathering, together with other sensors deployed at the environment (e.g., Internet of Things), contextual data from the environment and feedback from the workers. 

The remaining part of this work is structured as follows:
Section \ref{sec:2} describes a knowledge graph core methodology, Section \ref{sec:3} gives a brief background to fuzzy concept analysis and metric performance evaluation, Section \ref{sec:4} describes a simple scenario of Ticket Management and gives a short running example based on aircraft ticket dataset, Section \ref{sec:5} outlines the conclusion of this work and future directions. 


\begin{figure*}[h]
\centering
\includegraphics[width=0.8\linewidth]{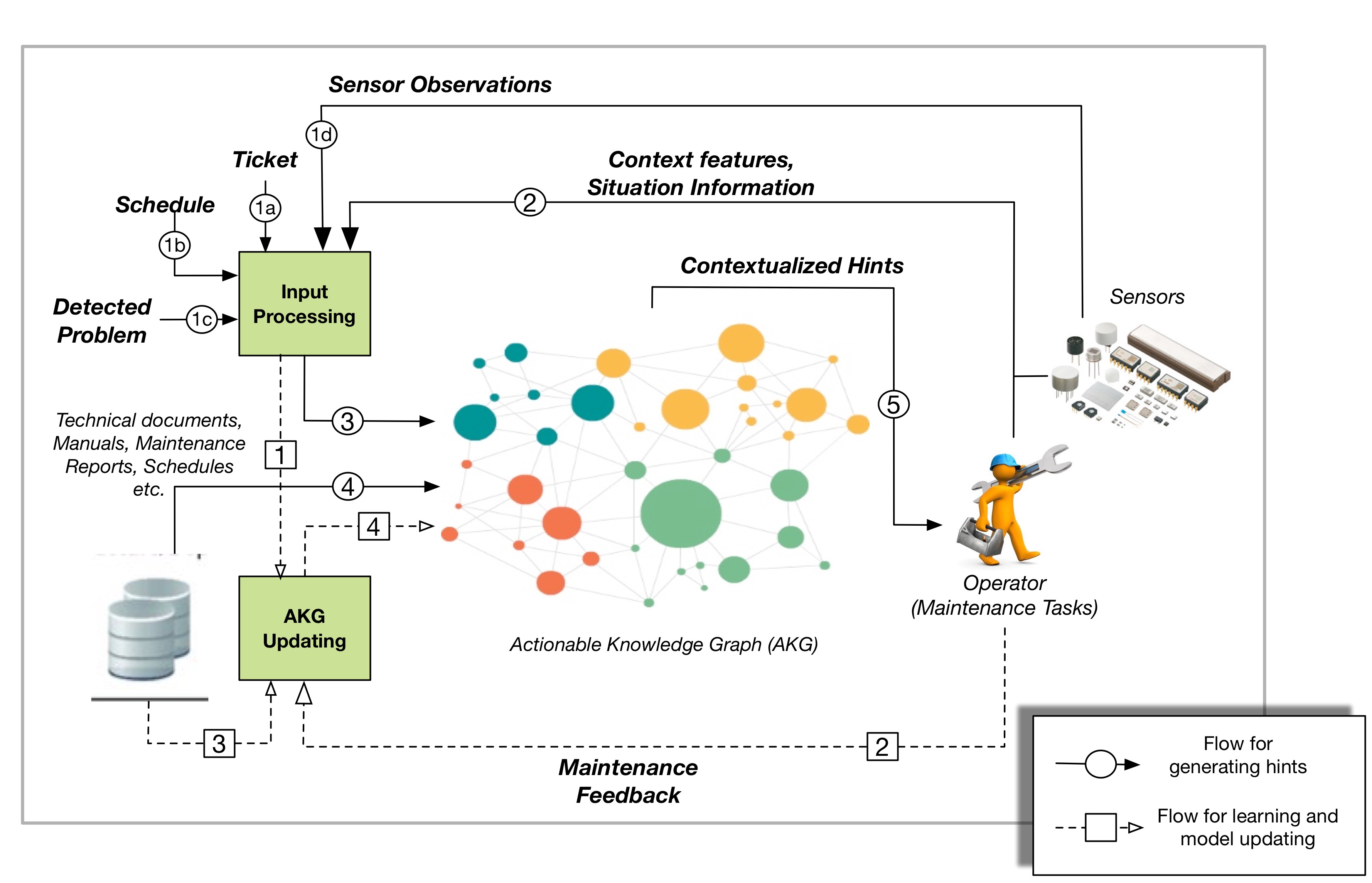}
\caption{Graphical representation of the overall approach.}
\label{oa}
\end{figure*}


\section{The Core Methodology}
\label{sec:2}
The core methodology of the proposed solution relies on the so-called Actionable Knowledge Graph (briefly, AKG) built by performing conceptual data analysis. Data refers to the database where the system keeps track of the events happening in the company during daily work activities, such as a ticket raised by the customer for a specific configuration, closure of a maintenance intervention, activities scheduling, sold configuration, and so on. Database transactions may be used to prepare a Formal Context and for executing the Formal Concept Analysis to carry out the lattice. The lattice is a Direct Acyclic Graph (DAG) in which each node (i.e., concept) includes two components, that are: attributes and objects, a.k.a. intent and extent, respectively. 
The intent is composed of the distinguishing features that characterize objects grouped in the same node. 
The extent is composed of the objects sharing the same set of features. 
These objects may be heterogeneous, such as tickets, tools, customers, configurations, and so on. 
The cardinality of each node may be used as analytics expressing, for instance, how many times tickets of the same type (sharing the same set/subset of attributes) are associated with the same maintenance  procedure, tools, and so on. 
In this sense, the attributes allow us to relate different items of heterogeneous nature in the resulting lattice, which is our AKG. 
In a nutshell, lattice nodes carry out how much an association is supported in the overall enterprise data center. Consequently, it is possible to provide useful insights by browsing the concepts of the AKG according to the features switched on during daily work activities.

Maintenance management typically falls into four main categories~\cite{10}: 
\begin{enumerate}
\item \emph{Reactive maintenance} involves repairing or remediating physical parts, components, or equipment only after they have broken down or been run to the point of failure.
\item \emph{Planned maintenance} consists of replacing parts, components or equipment before they fail, and, being time-based preventative maintenance, it can help avoid broken machinery and decrease downtime.
\item \emph{Proactive maintenance} aims at identifying and addressing the problems that can lead to machine failures.
\item \emph{Predictive maintenance} has the objective to predict data when and where failures could occur by gathering data from connected, smart machines and equipment and potentially allows to maximize the efficiency of the interventions and to minimize unnecessary downtime.
\end{enumerate}

The methodology we are proposing can support all the previous four maintenance strategies (which are enumerated in order of increasing complexity) by opportunely adjusting attributes and objects belonging to formal context. In particular, it is possible to configure the set of attributes of the formal context based on the adopted strategy. For instance, for handling reactive maintenance, it is needed to consider some attributes representing information about failures; for planned maintenance, it is required to insert time-related information as one or more attributes; in the case of proactive maintenance, it is possible to model failure causes as attributes and, lastly, if predictive maintenance is the adopted strategy, it is needed to represent sensor observations as attributes.

Fig. \ref{oa} shows two different execution flows, having the AKG as the computation core. 
The first one is the \emph{generation of hints} that starts from a specific triggering event (step 1a, 1b, 1c, 1d), gathers all input data, including context features and situation information (step 2), and executes a query over the AKG (step 3) to obtain and deliver (step 5) a ranked list of hints containing digital content (retrieved during step 4) for supporting worker during a maintenance operation. The triggering event could be a ticket (1a), a timer signal from a schedule (1b), the detection of a specific problem (1c), or the occurrence of specific sensor observations (1d). Each one of the above triggering events could be useful to implement one of the aforementioned maintenance strategies.  
The second flow is the \emph{learning and model updating} phase in which the AKG is updated incrementally by using traces coming from several sources. In particular, it is possible, for the AKG, to learn from the operator's feedback. Such feedback is correlated to context information, sensor observations, triggering events, and other entities to refine the model represented by the AKG.

The following section introduces the theory of Formal Concept Analysis on which relies the graph building background; we will describe how the graph is built. 

\section{Actionable Knowledge Graph}
\label{sec:3}

Digital resources available in the company are preprocessed and vectorized according to their features. 
Then they are collected and processed in batch to build the initial knowledge graph; successively, the graph is exploited to carry out links and suggestions for supporting daily work activities.
In addition, it is necessary to include a mechanism capable of updating the knowledge graph according to the feedback provided by the employee (the user, the operator) by selecting or discarding the proposed hints.

\subsection{Data Gathering and Preprocessing}
Available digital resources (e.g., problem-and-incident tickets) are collected and preprocessed to extract significant aspects and features. In particular, by means of domain taxonomies (e.g., \emph{ATA 100}\footnote{$https://av-info.faa.gov/sdrx/documents/JASC_Code.pdf$} for aircraft classification) and a specific Named Entity Recognizer (NER), the main keywords are determined. Keywords, together with equipment and context information, become features for the graph building process.  As expressed in the following section, they will populate the formal context as attributes. Context information regards the event's time and location, involved people, etc., that also contribute to the attribute set definition (see Section \ref{sec:example} for a practical example).

\subsection{Building}

The formal model behind the proposed methodology is the fuzzy extension of Formal Concept Analysis (briefly, Fuzzy FCA, or FFCA) \cite{4}. FCA  is a theoretical framework that supplies a basis for conceptual data analysis, knowledge processing, and extraction. Fuzzy FCA \cite{5} combines fuzzy logic into FCA, representing the uncertainty through membership values in the range [0, 1]. In particular, 
Fuzzy FCA deals with fuzzy relations between objects 
and their features considering membership varying in [0,1], instead of the binary relation of traditional FCA. So it enables us to specify more or less relevant features to represent resources enabling the granular representation of them and to carry out similarity among resources, varying in [0,1].  

Following, some definitions about Fuzzy FCA are given.

\begin{definition}\label{def1}
	{\it A {\bf Fuzzy Formal Context} is a triple $K=(G,M,I)$, where G is a set of objects, M is a set of attributes, and $I= \left ( (G \times M),\mu \right )$ is a fuzzy set. }
\end{definition}
Recall that, being $I$ a fuzzy set, each  pair $(g, m)\in I$ has a membership value $\mu(g, m)$  in [0,1]. In the following, the fuzzy set function $\mu$ will be denoted by $\mu_I$.

\begin{figure*}[!h]
	\centering
	\caption{An example of Fuzzy Formal Context and corresponding Fuzzy Concept Lattice, it is used also as sample scenario in Section \ref{sec:example}. }
	\includegraphics[width=0.7\linewidth]{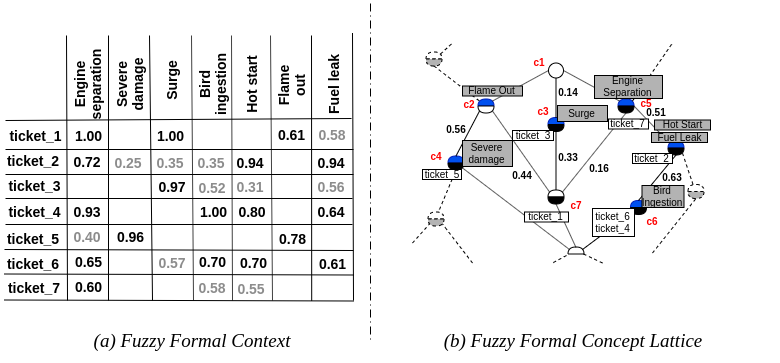}
	\label{ctx}
\end{figure*}

\begin{definition}	{\bf Fuzzy Representation of Object}. { \it Each object O in a fuzzy formal context $K$ can be represented by a fuzzy set $\Phi(O)$ as $\Phi(O)$=\{A$_1(\mu_1)$, A$_2(\mu_2)$,\dots, A$_m(\mu_m)$\}, where \{A$_1$, A$_2$,\dots, A$_m$\} is the set of attributes in $K$ and $\mu_i$ is the membership of O with attribute A$_i$ in $K$. $\Phi(O)$ is called the fuzzy representation of O.}
\end{definition}
The Fuzzy Formal Context (see Definition \ref{def1}) is often represented as a cross-table, as shown in Fig. \ref{ctx} (a), where the rows represent the objects, while the columns, the attributes. Each cell of the table contains a membership value in [0, 1], but the Fuzzy Formal Context showed in Fig.  \ref{ctx} (a)  has a confidence threshold $T=0.6$, that means all the relationships with membership values less than 0.6 are not shown. 

Taking into account Fuzzy Formal Context, the Fuzzy FCA algorithm is able to identify Fuzzy Formal Concepts and subsumption relations among them. More formally, the definition of Fuzzy Formal Concept and order relation among them are given as follows:

Given a fuzzy formal context $ K=(G, M, I)$ and a confidence threshold $\chi$, for ${G}' \subseteq G$  and ${M}'\subseteq M$, we define $G^{*}=\{m\in M\ |\ \forall g \in G',\ \mu_I(g,  m)\geq  \chi\}$ and $M^{*}=\{g \in G \ |\ \forall m \in {M}',\ \mu_I(g, m) \geq \chi\}$.

\begin{definition}\label{mu}{\bf Fuzzy Formal Concept}. {\it A  fuzzy  formal  concept  (or fuzzy  concept) $C$ of  a  fuzzy  formal  context  K  with  a  confidence  threshold $\chi$, is $C=(I_{G'},M')$, where, for $G'\subseteq G,\ \  I_{{G}'} = \left(G', \mu \right), {M}'\subseteq M, {G}^*={M}'$  and ${M}^*={G}'$. Each object g has a membership $\mu_{{I_{G'}}}$ defined as}
	\begin{equation}
	\mu_{{I_{G'}}}(g)= min_{m \in {M}'} \left (\mu_{I}(g, m)\right )
	\end{equation}
	\noindent {\it where $\mu_{I}$  is the fuzzy function of $I$.}\\
\end{definition}
Note that if $M'=\emptyset$ then $\mu_{I}(g)= 1$ for every $g$. $G'$ and $M'$ are the extent and intent of the formal concept $\left (I_{{G}'}, {M}'\right )$, respectively.


\begin{definition}{\it Let $(I_{{G'}}, M')$ and $(I_{{G''}}, M'')$ be two fuzzy concepts of a Fuzzy Formal Context $(G, M, I)$.  $(I_{{G'}}, M')$ is  the  \textbf{sub-concept}  of $(I_{{G''}}, M'')$,  denoted as $(I_{{G'}}, M')\leq( I_{{G''}}, M'')$,  if  and  only  if  $I_{{G'}}\sqsubseteq I_{{G''}} (\Leftrightarrow M'' \subseteq M')$. Equivalently, $(I_{{G''}}, M'')$ is the \textbf{super-concept} of $(I_{{G'}}, M')$.}
\end{definition}

For instance, let us observe in Fig. \ref{ctx} (b), the concept $c_2$ is a \textit{sub-concept} of the concept $c_1$. Equivalently the concept $c_1$ is a \textit{super-concept} of the concept $c_2$. 
Let us note that each node (i.e., a formal concept) is composed of the objects and the associated set of attributes emphasizing, through the fuzzy membership, the objects that are better represented by a set of attributes. In Fig. \ref{ctx} (b), each node can be colored differently, according to its characteristics: a half-blue colored node represents a concept with $own$ attributes; a half-black colored node instead, outlines the presence of $own$ objects in the concept; finally, a half-white colored node can represent a concept with no $own$ objects (if the white-colored portion is the half below of the circle) or attributes (if the white half is up on the circle). 
Furthermore, given a Fuzzy Formal Concept of Fuzzy Formal Context, it is easy to see that the sub-concept relation $\leq$ induces a \textit{Fuzzy Lattice} of Fuzzy Formal Concepts. The lowest concept contains all attributes, 
and the uppermost concept contains all objects 
of Fuzzy Formal Context.

In addition, the notion of \emph{Fuzzy Formal Concept Support} (briefly, \emph{Support}) allows us to measure how much frequently objects belonging to the same node of the lattice are associated. The \emph{Support} is used to carry out statistical information supporting system suggestions.
The notion of Support is based on the definition of \emph{frequent concept intent} and closure systems introduced in \cite{11}. Specifically:

\emph{\bf Definition 6: }\textit{\textbf{Fuzzy Formal Concept Support}. Let $K=(G,M,I)$ be a fuzzy formal context, the support of a Fuzzy Formal Concept $C'=(I_{G'}, M')$ is given by}
\begin{equation}
Supp(C') = \frac{|G'|}{|G|}
\end{equation}

\noindent Let \emph{minsupp} be a threshold $\in [0-1]$, then $C'$ is said to be a frequent concept if $Supp(C')\geq minsupp$. 

The main drawback of FCA and FFCA is that they become prohibitively time consuming as the dataset size increases. Since there is a deluge of sensor data acquired through a sensor cloud architecture, the proposed framework exploits a MapReduce implementation of Formal Concept Analysis adopting Hadoop MapReduce by Apache\footnote{http://hadoop.apache.org/mapreduce/}\cite{12}.  


\subsection{Exploitation}\label{section:3.3}
The system assists the operator during daily work activities by providing recommendations and handling the feedbacks exploiting the resulting updated AKG extracted as described before. 
These suggestions are supplied by using a matching algorithm described below. The algorithm, given a set of features happening in a particular scenario, such as troubleshooting, execution of maintenance intervention, and so forth, ranks the hints according to the matching evaluation between these input features and the statistical information underlying the AKG.


Let us suppose that the system receives a set of features $F$ with values $F=\{f_{i_1},f_{i_2}, \dots, f_{i_m}\}$ and let us consider the $j-th$ lattice concept $C_j$ containing attributes $A_j=\{a_{i_1},a_{i_2}, \dots, a_{i_k}\}$ and objects $O_j=\{o_{i_1},o_{i_2}, \dots, o_{i_h}\}$ each of them has a membership belonging degree to the concept $C_j$, that is $\mu$.
Then, the matching degree is evaluated by performing F-Measure on the results of Precision $P_{i,j}$ and Recall $R_{i,j}$ as follows~\cite{16}:
\begin{equation}
P_{i,j}=\frac{\left | F\bigcap A_j\right |}{\left | A_j \right |} \ R_{i,j}=\frac{\left | F\bigcap A_j \right |}{\left | F \right |}
\end{equation}
\begin{equation}
\label{fmeasure}
F_{i,j}=2\times \frac{P_{i,j}\times R_{i,j}}{P_{i,j}+R_{i,j}}  
\end{equation}
The evaluation of the intersection between $A_j$ and $F$ is computed as the maximum cardinality bipartite matching considering the graph $G=\left \langle V, E \right \rangle$:  \\ 
\begin{equation}
V =\left \{ F \bigcup A_j \right \}
\end{equation}
\begin{equation}
\begin{matrix}
E=\{ (x,y) \ | \ x \in F \ and \ y\in A_j\ if \ rel(x,y) \geq 0.7\} 
\end{matrix}
\end{equation}
and $rel(x,y)$ is a relatedness value expressing a partial matching between feature and attribute, i.e., a similarity value (e.g., Wikipedia Linked Measure \cite{7}). The partial match allows us to represent a greater set of situations that may happen in the real scenario, a more generalized AKG. 

The results of matching degree combined with the concept belonging degree of the objects in $O_j$ to the concept $C_j$, i.e., the membership function $\mu$, are exploited to rank the resulting objects, that are given as output recommendations. 
Let us note that the matching degree is evaluated with the overall set of concepts in AKG by following a top-down visit from the root to the target concept. 

\begin{table*}[t]
\caption{Troubleshooting aircraft dataset sample}
\centering
\begin{tabular}{|c|c|c|c|c|}
\hline
\textbf{Model}   & \textbf{Selling Country} & \textbf{Selling Year} & \textbf{Client} & \textbf{Symptoms' Description} \\ \hline
Boeing 777-300ER & USA                      & 2019                  & Emirates        & Reverser inadverted deploy                       \\ \hline
Boeing 777-9X    & USA                      & 2018                  & Emirates        & Fuel leak, Engine separation                      \\ \hline
A380-800         & Germany                  & 2018                  & EasyJet         & Engine separation                        \\ \hline
Boeing 777       & Italy                    & 2019                  & Emirates        & Hot start, Engine separation                      \\ \hline
A330-200         & France                   & 2017                  & Alitalia        & Tail pipe fires                    \\ \hline
\end{tabular}
\label{tab:dataset}
\end{table*}

\begin{table*}[t]
\caption{Lattice concepts with corresponding F-Measure}
\centering
\begin{tabular}{|c|c|c|}
\hline
\textbf{Concept}   & \textbf{Attributes} & \textbf{F-Measure} \\ \hline
\textbf{$c_6$}        &    \{EngineSeparation, HotStart, FuelLeak, Birdingestion\}   &   \textbf{$0.85$}  \\ \hline
$c_5$       &    \{EngineSeparation\}   & $0.5$    \\ \hline
$c_7$        &    \{EngineSeparation, Surge\}   &  $0.5$   \\ \hline
\end{tabular}
\label{tab:results}
\end{table*}

\section{Sample scenario: Tickets Management}
\label{sec:4}
Ticket management may be really improved by introducing an intelligent mechanism to collect and reuse enterprise knowledge. 
In this sense, AKG allows us to cross-relate similar solutions for addressing similar problems; it provides useful hints by considering the context in which the question described in the ticket is raised, for instance, about the equipment to use for executing maintenance operation for an installed configuration, and so forth.
Besides, the AKG may provide useful insights for planning the activities taking into account the skills required by the tickets and the currently available ones in the organization. 
Moreover, the availability of a rich set of logs coming from a smart environment, especially in the Industry 4.0 perspective,  storing each session of work performed for solving the tickets could be very useful for also implementing predictive maintenance process, automating maintenance orders and generating maintenance tickets automatically. 

The hits providing services realized on the top of the proposed AKG have been tested for addressing ticket management. 
In particular, the data coming from the ticket received by the customer are used for providing features to select the suggestions on the AKG. By taking advantage of AKG, it is also possible to configure faceted browsing~\cite{16} guided from ticket features, in order to simplify the navigation of data and identify target solutions easily. By applying the filters, called facets, users can narrow down search results, in different domains \cite{17}.

In a usual scenario, the ticket contains at least the following information:
\begin{itemize}
	\item customer info;
	\item problem description;
	\item configuration;
	\item timestamp.
\end{itemize}

This information is used by the system, for instance, to retrieve the configuration characterizing the product sold to the customer, its age and its current state in the product life cycle, the last performed intervention on the item, and so forth. In addition, the problem description is processed along with a pipeline task for text mining; the resulting concepts are used as additional filters on the AKG to select the right hints to provide.
The prepossessing aims to get the features that allow activating the search in the AKG.
The evaluation has been performed in terms of the average time for solving tickets and the number of tickets that have been addressed last year with respect to the previous year.

\subsection{Running Example}\label{sec:example}
In this section, we will build a running example applying the defined workflow to a simple scenario of ticket management, introduced in Section \ref{sec:4}.
Let us suppose having the troubleshooting aircraft dataset shown in Table~\ref{tab:dataset}. It contains features characterizing tickets, like aircraft model, customer details, symptoms, and so forth. 

When a new ticket is received, the system will search in the Actionable Knowledge Graph obtaining a set of possible suitable solutions applied to solve similar problems in the past. 
The system will select the tickets most similar to the new one and the corresponding solutions by using the method described in Section \ref{section:3.3}. 
For example, let us suppose to be aircraft seller and to receive a ticket T defined as follows:
\begin{equation}
T = (Boeing 777-300ER, USA, Emirates, S)
\end{equation}
where $S=\{EngineSeparation, HotStart, FuelLeak\}$ are all malfunctioning symptoms that the client has found at the running time. When T arrives, we search for the best concepts in our knowledge graph (showed in Fig. \ref{ctx}) by calculating ones having the highest F-Measure. Basing on F-Measures listed in Table \ref{tab:results}, the concept with the highest value is $c_6$. So, the suggestions will regard tickets falling in concept $c_6$ (i.e., $ticket_6$ and $ticket_4$).

Finally, the used solutions will be included in the knowledge base of the company allowing the implementation of incremental learning.

\section{Conclusion}
\label{sec:5}

This paper proposes a solution based on FFCA for constructing AKG. The idea is to collect heterogeneous historical information (i.e., from manuals, documents, and context) and group it by means of the fuzzy lattice. 
As a practical example, the proposal shows how the FFCA is a powerful tool for retrieving and recommend useful hints for supporting daily maintenance activities. In general, the construction of a so-called Actionable Knowledge Graph starting from structured and unstructured content may support many more scenarios.
Through AKG, it has been possible to structure poorly structured information through the use of the knowledge graph and easily consulting it, for instance, by means of faceted browsing of the retrieved suggestions. 
Since this work showed a proof of concept, in the future, it could be interesting extending the method by presenting exhaustive experimentation. A possible extension is considering process parameters coming, for instance, from wearable device or wearable sensor exploiting technologies such as Augmented or Mixed reality for data visualization.

\section*{Acknowledgment}
This research was partially supported by the ECSEL-JU under the program ECSEL-Innovation Actions-2018 (ECSEL-IA) for research project CPS4EU (ID-826276) research project in the area Cyber-Physical Systems.

 \bibliographystyle{IEEEtran}
 \bibliography{mybib}{}
 
\end{document}